%% file: root.tex
\documentclass[letterpaper, 10 pt, conference]{ieeeconf}  

\IEEEoverridecommandlockouts                              

\usepackage{caption}
\usepackage{hyperref}
\usepackage{graphicx}
\usepackage{amsmath} 
\usepackage{amssymb}  

\usepackage{algpseudocode}
\usepackage{algorithm}
\usepackage{booktabs}
\usepackage{multirow}

\usepackage{lipsum}
\usepackage{etoolbox}
\usepackage{caption}
\usepackage{titlesec}

\newcommand{\shrink}{\def\baselinestretch{0.99}\large\normalsize} 
\shrink

\title{\LARGE \bf
Latent Policy Steering: An Efficient and Flexible Framework for Cross-Embodiment Transfer
}

\author{Yiqi Wang$^{1}$ \thanks{$^{1}$Authors are from Robotics Institute, School of Computer Science, Carnegie Mellon University, 5000 Forbes Ave, Pittsburgh, United States {\tt\small yiqiw2@andrew.cmu.edu}}  and Mrinal Verghese$^{1}$ and Jeff Schneider$^{1}$}

\makeatletter
\apptocmd{\@maketitle}{\centering\input{Title_Figure}}{}{}

\makeatother

\begin{document}

\maketitle
\thispagestyle{empty}
\pagestyle{empty}

\begin{abstract}

The performance of learned robot visuomotor policies depends heavily on the size and quality of their training data, yet collecting high-quality demonstrations remains costly for robots in the real world. Although large-scale robot and human datasets are increasingly available, embodiment gaps and mismatched action spaces make them difficult to leverage directly. Cross-embodiment transfer, reusing experience from other embodiments to improve learning on a target embodiment, is therefore crucial for scaling robot learning beyond per-robot data collection.

In this work, we find that efficient transfer can be achieved by learning from what is shared across embodiments, the visual dynamics of how the world responds to motion, and by effectively exploiting the scarce target-embodiment data at test time. The proposed framework, called Latent Policy Steering (LPS), implements an embodiment-agnostic pretraining phase, which trains an image-based World Model (WM) with optical flow across diverse embodiments. The resulting WM is finetuned on the target embodiment with robot actions. It then steers the base policy toward better actions by searching in the WM's latent space for plans that stay close to the finetuning data. LPS is a policy-agnostic framework: it can flexibly accommodate different policies without having to retrain them. In Robomimic and real-world evaluations, LPS improves the average performance of Diffusion Policy relatively by 16\% and 62\%, and Pi0.5 by 8\% and 14\%, with only 50 demonstrations on an unseen target embodiment.


\end{abstract}

\section{INTRODUCTION}

Imitation learning through Behavior Cloning (BC) is a widely adopted paradigm to acquire visuomotor policies for robots \cite{brohan2022rt}\cite{chi2023diffusion}\cite{shafiullah2022behavior}. To achieve high task success, sufficient expert demonstrations must be collected, which is a time-consuming process. Furthermore, the data collected are often specific to a robot, a task, or an environment, making them difficult to apply directly to other embodiments or environments. By collecting large datasets across different robots and environments \cite{vuong2023open}, progress has been made toward building generalist robot policies via cross-embodiment training \cite{team2024octo}\cite{henighan2020scaling}.  However, these models sometimes do not generalize well to new robots or tasks. Finetuning them for better performance requires a considerable amount of data, given the large model sizes needed to learn from large datasets \cite{brohan2022rt}\cite{black2410pi0}. For example, $\pi_0$ \cite{black2410pi0} requires 5--10 extra hours of finetuning data to achieve high success rates on new tasks.

In typical LLM/VLM practice, a model trained on large and diverse datasets across multiple tasks will produce general representations that are transferable to new tasks with only a small amount of data, known as pretraining \cite{henighan2020scaling}\cite{kaplan2020scaling}. However, learning transferable representations to reduce robot data collection is particularly challenging: embodiment gaps arising from mismatched action spaces and proprioception across multi-robot datasets can lead to embodiment-dependent representations that transfer poorly to a new embodiment.

To achieve more efficient cross-embodiment transfer, recent efforts focus on what embodiments share rather than on what sets them apart, so that diverse, existing experience, including public robot data and cheap data from very different embodiments such as humans, can benefit a new robot. 
A common approach \cite{team2024octo}\cite{wang2024scaling} extracts shared representations with a common trunk while handling embodiment-dependent modalities with ad-hoc stems and heads. More recently, other works propose learning a shared latent action space from visual observations to pretrain VLAs \cite{ye2025latent}\cite{chen2025moto} or WMs \cite{gao2026dreamdojo}, where the latent actions are replaced by robot actions during finetuning. However, the learned latent actions are not guaranteed to be meaningful and require additional training. 

\begin{figure}[t]
      \centering
     
      \includegraphics[scale=0.042]{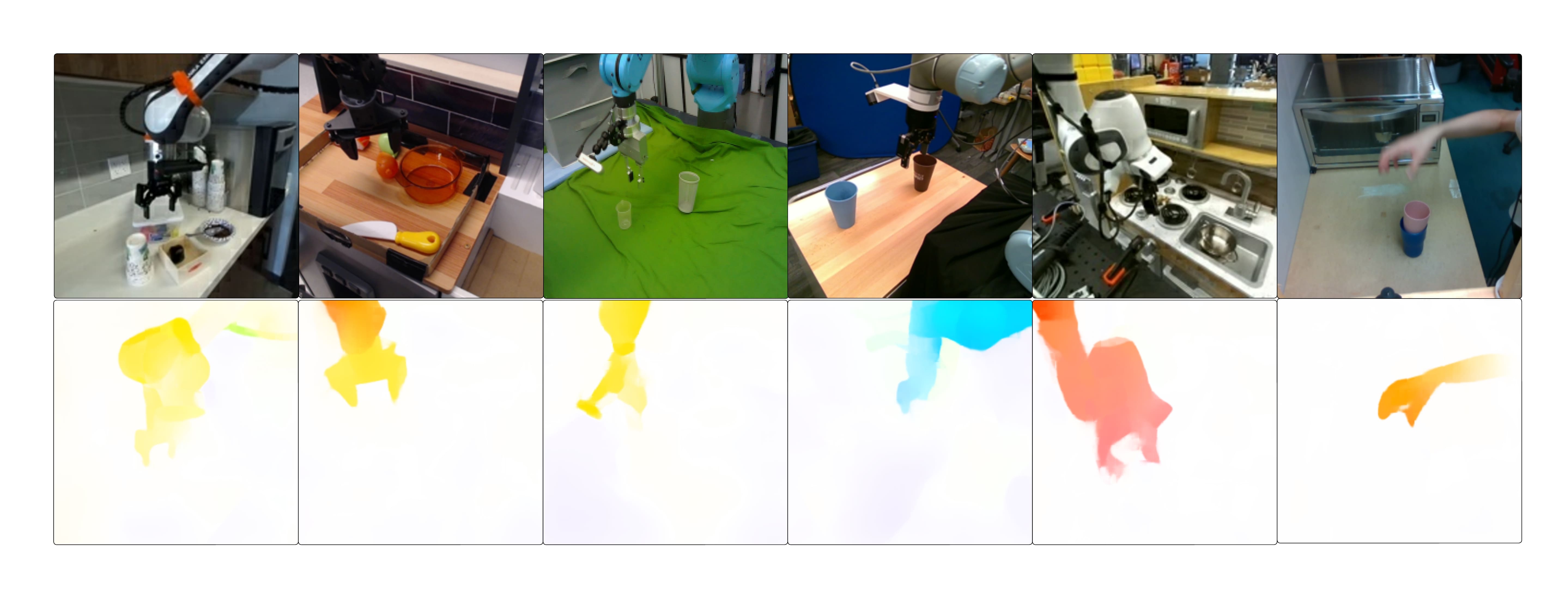}
      \caption{ 
      \textbf{Optical flow reduces embodiment gaps}. We observe that motions captured by optical flow across embodiments are similar in visual space. For example, picking up a cup will generate similar flow vectors in both the agent and the cup, regardless of the agent embodiment. By using optical flow as a transition representation of visual dynamics, we reduce the pretrained model's dependency on specific embodiments, improving transferability to a new embodiment. }
      \vspace{-1em}
      \label{fig:optical_flow_visualization}
   \end{figure}

Instead of inferring an action space from visual observation, previous work \cite{lin_flowretrieval_2024}\cite{bi2026motus} notes that \textit{the skills performed across different embodiments produce a visually similar motion}, which can be captured by \textit{off-the-shelf representations such as optical flow}. Inspired by the observation that optical flow is widely available across embodiments and carries structured information about visual motion (as shown in Fig.~\ref{fig:optical_flow_visualization}), we propose an \textit{embodiment-agnostic pretraining} stage that obtains transferable representations from this off-the-shelf signal: an image-based WM (Dreamer \cite{hafner2019learning}\cite{hafner2023mastering}) is pretrained across various embodiments with encoded optical flow (left and middle of Fig.~\ref{fig:motivation1}). In contrast to previous work \cite{lin_flowretrieval_2024}\cite{bi2026motus}, which learns to encode optical flow in a separate stage before pretraining and thus incurs additional training cost, we use a convolution-based encoder that is trained end-to-end with the WM and produces a compact optical flow representation that suppresses motion-irrelevant information. 

Another observation we make is that the existing practices for cross-embodiment transfer are highly heterogeneous, due to different design choices of the policy, such as specialists \cite{pace2025x}\cite{bauer2025latent}, generalists built from VLMs \cite{black2410pi0}\cite{brohan2023rt}\cite{intelligence2025pi}\cite{kim_openvla_2024}, and multi-modal architectures \cite{bi2026motus}\cite{lyu2026lda}. Thus, to accommodate differences in the policy during cross-embodiment transfer, we propose a \textit{policy-agnostic finetuning} stage (middle and right of Fig.~\ref{fig:motivation1}): the WM is finetuned with robot actions on the target embodiment. The WM and the policy can then be combined, regardless of the choice of policy, to simulate states likely to be visited during inference and to train a value function that penalizes distribution shift away from the dataset. During inference, the WM and the value function evaluate multiple action plans sampled from the policy and select the best one for execution, often referred to as Policy Steering \cite{nakamoto2024steering}\cite{wang2024inference}\cite{wu2025foresight}. As a result, the robot is robust to inference-time distribution shift, since it has been simulated during finetuning, addressing a prevalent problem for sequential decision-making \cite{fujimoto2019off}\cite{yu2020mopo}\cite{kumar2020conservative}. 

We refer to our proposed framework for cross-embodiment transfer as Latent Policy Steering (LPS). Our contributions are as follows.
\begin{itemize}
    \item We propose to pretrain an embodiment-agnostic World Model (WM) across different robots and human embodiments, leveraging optical flow as an off-the-shelf, embodiment-agnostic transition representation. 
    \item We develop a policy-agnostic finetuning phase: the WM can flexibly accommodate different policies (e.g., Diffusion Policy \cite{chi2023diffusion}, state-of-the-art VLAs \cite{intelligence2025pi}) to improve their inference performance by searching for better actions in the WM's latent space. 
    \item Given 50 demonstrations provided on an unseen target embodiment, LPS improves the average performance of the Diffusion Policy relatively by 16\% and 62\%, and Pi0.5 by 8\% and 14\%, in the Robomimic and real-world evaluations, demonstrating the efficiency of the proposed framework in the low-data regime.
\end{itemize}

\input{related_work}    

\input{method}

\input{experiment}

\section{CONCLUSIONS}
We develop an efficient and flexible framework for cross-embodiment transfer, called Latent Policy Steering (LPS). Through embodiment-agnostic pretraining, LPS transfers existing and cost-effective data sources such as public multi-embodiment robot datasets \cite{vuong2023open}, robot data from simulations \cite{robosuite2020}, and easily collected human data from play to an unseen target embodiment with only 50 demonstrations. LPS's performance scales effectively when provided with more pretraining data from diverse embodiments, and it substantially improves over the base policy when the target-embodiment dataset is small. Thanks to the policy-agnostic finetuning phase, LPS can accommodate different base policy choices, such as a lightweight Diffusion Policy that learns from scratch or a state-of-the-art VLA, by simulating the test-time distribution shift in the World Model's latent state space and steering the policy to stay close to its training data during inference. We hope that LPS demonstrates a paradigm complementary to ongoing research on cross-embodiment transfer, further lowering the deployment cost of data-driven robot learning.

\section{DISCUSSION}

We realize that existing practices that conduct cross-embodiment transfer via building VLAs from VLMs (e.g., Pi0.5) remain very effective: with sufficient model capacity and a large amount of data, VLAs extract transferable representations from multi-embodiment datasets without explicitly mitigating embodiment gaps and outperform LPS on simpler tasks that involve pick-and-place with common objects. However, pretraining VLAs requires aggregating datasets with action labels, which could be costly to obtain. On the other hand, LPS takes advantage of off-the-shelf transition representations such as optical flow, which turn label-free data (e.g., human videos) into pretraining data, reducing embodiment gaps across diverse embodiments without incurring the additional training phase required by the latent action approaches \cite{ye2025latent}\cite{gao2026dreamdojo}\cite{bi2026motus}.

Although optical flow enables LPS to leverage cost-effective label-free data sources, it has several limitations. First, optical flow cannot reliably capture motion under occlusion. Furthermore, optical flow is viewpoint-dependent. Given different viewpoints, the same skill results in different optical flow patterns, making learning visual dynamics more difficult. However, it is possible to overcome this limitation by leveraging large cross-embodiment datasets with diverse viewpoints. Additionally, Internet-scale videos with extensive camera motion produce noisy optical flow, which remains unexplored in this work. In the future, we plan to further investigate the benefits of leveraging off-the-shelf representations such as optical flow to improve cross-embodiment transfer at scale.

\begin{figure}[t]
  \centering
  \includegraphics[width=0.86\columnwidth]{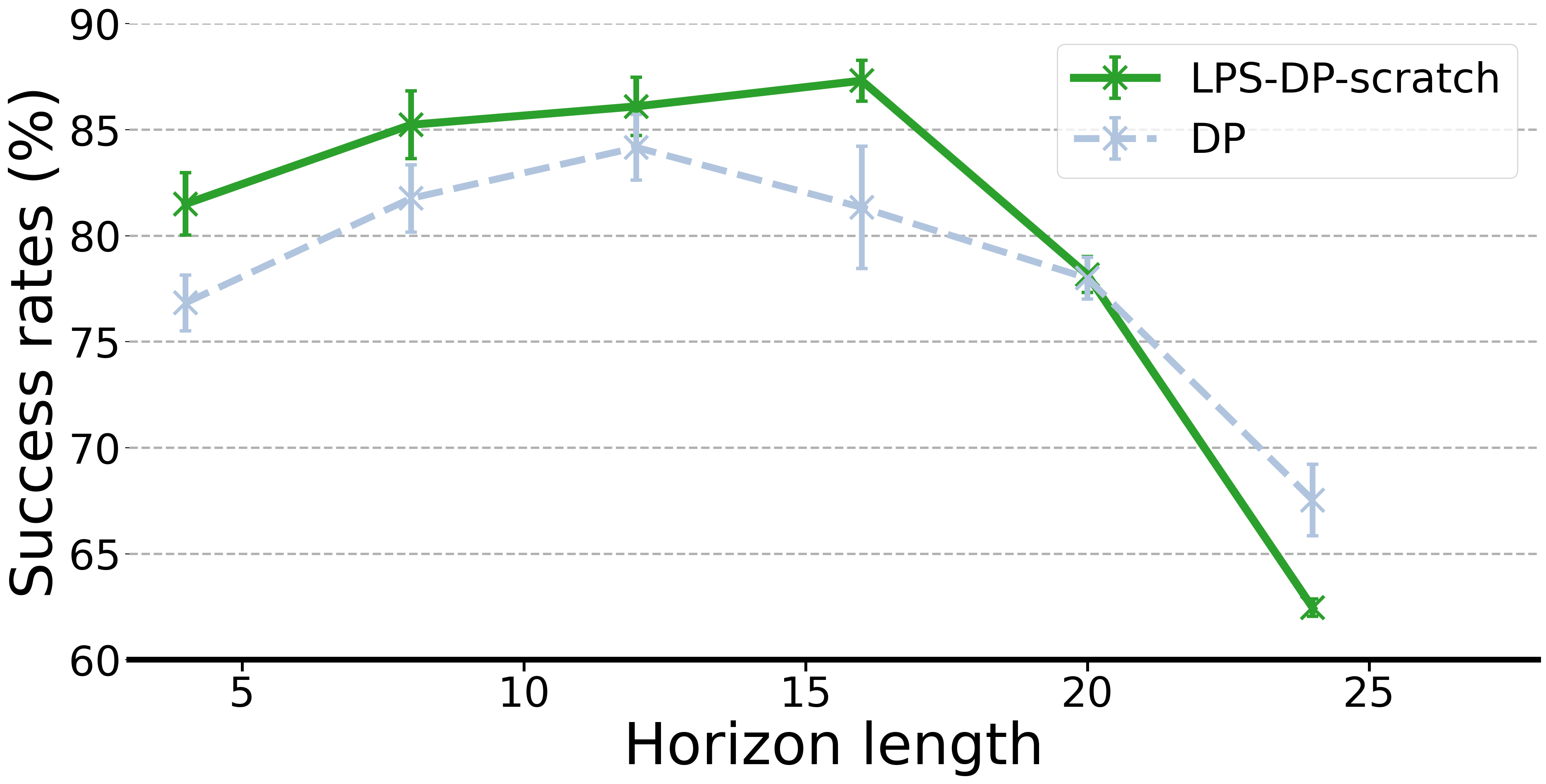}
  \caption{ \textbf{Effective horizons.} LPS effectively improves policy performance for horizons up to length 16. When the horizon becomes very long (e.g., 20, 24), the reward used to capture distribution shift becomes noisy, making the value function less effective during inference. }\label{fig:lps_horizon}
  \vspace{-2em}
\end{figure}









\bibliographystyle{IEEEtran}
\bibliography{root}

\end{document}

%% file: Title_Figure.tex
\includegraphics[width=.99\linewidth]{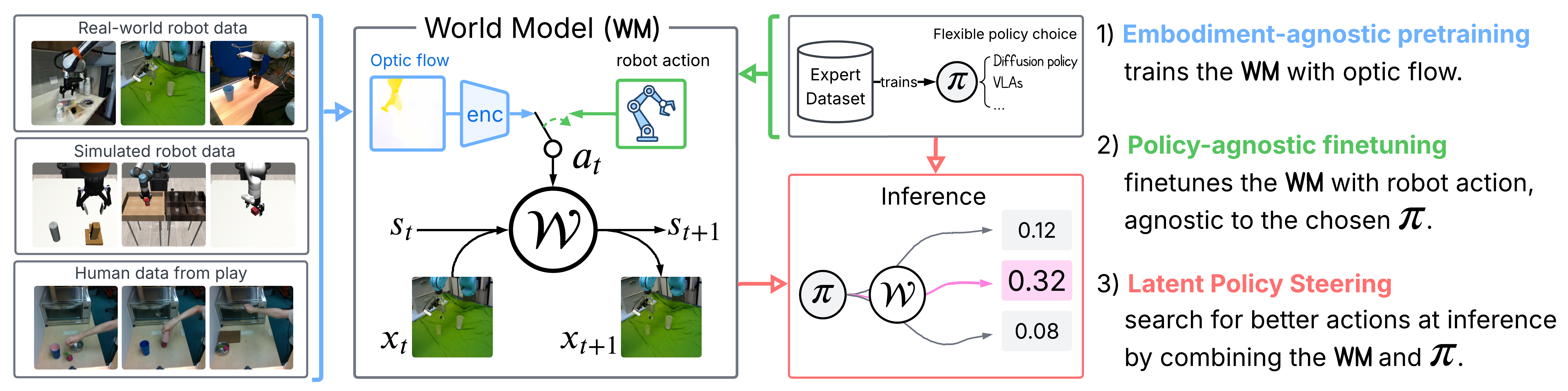}
\captionof{figure}{
\textbf{Overview}: 1) An image-based World Model (WM), which learns visual dynamics conditioned on optical flow as embodiment-agnostic transition representations, is pretrained across diverse embodiments. 2) After the WM is finetuned on the target embodiment conditioned on robot actions, it can be combined with different policies to achieve better performance at inference by 3) sampling multiple candidates from the policy and executing the best one.
}
\setcounter{figure}{1}
\label{fig:motivation1}
\vspace{-3mm}

%% file: related_work.tex
\section{Related Work}

\subsection{Robot Learning from Diverse Data}
Collecting large amounts of robot data for a specific task can be challenging and time-consuming. A common strategy is to leverage data from other sources, including public robot datasets with multiple types of robots and even human videos. Prior works \cite{karamcheti_language-driven_2023}\cite{nair2022r3m}\cite{majumdar_where_2024}\cite{radosavovic2023real} train a visual encoder using both human and robot video data to avoid learning the policy network from scratch. However, such a pretrained encoder only provides visual representations and does not itself learn to make decisions. Thus, recent work has also looked at learning a single policy network across multiple robot embodiments using a modular structure with a common trunk and separate heads for different action spaces \cite{team2024octo}\cite{wang2024scaling}. One popular paradigm is the Vision-Language-Action model (VLA) approach, where a pretrained Vision-Language Model (VLM) is finetuned with robot data to act as a robot policy \cite{brohan2023rt}\cite{black2410pi0}\cite{kim_openvla_2024}. To reduce embodiment gaps during pretraining, recent work \cite{bi2026motus}\cite{chen2025moto} has explored training VLMs or WMs \cite{gao2026dreamdojo}\cite{gao2025adaworld} with latent action labels \cite{ye2025latent} shared across embodiments via a VQ-VAE \cite{van2017neural}, then finetuning on robot actions to obtain VLAs. We compare our approach with a finetuned HPT \cite{wang2024scaling} and Pi0.5 \cite{intelligence2025pi}, as representative examples of the modular and VLA approaches.

\vspace{-0.1em}

\subsection{Policy Steering \& Planning with World Models} 

World Models (WMs) based on the Recurrent State-Space Model (RSSM) have become an increasingly common approach for modeling environment dynamics and transition functions in robotics. Popularized by Hafner et al. \cite{hafner2019learning}\cite{hafner2023mastering}, these models consist of an encoder that maps visual observations to a latent state representation, a transition function that predicts future latent states, and a decoder that is used primarily during training to propagate gradients. Once trained, such a WM can be utilized in a variety of ways. For example, one could learn actor and critic models based on the simulated data from the WM \cite{hafner2023mastering}. Another approach is to encode a goal image in the latent space and then use the similarity between planned states and the goal image as a value function to select the best sequence of actions. These action sequences are commonly optimized using gradient-free optimizers such as the Cross-Entropy Method (CEM), but gradient-based optimization can also be used~\cite{zhou2024dino}\cite{mendonca2023structured}\cite{hafner2023mastering}.

These aforementioned approaches are closely related to policy steering, where a value function and optionally a WM are used to refine the output of a base policy~\cite{nakamoto2024steering}\cite{wu2025foresight}. Policy steering has been shown to compensate for failure modes in the base policy~\cite{nakamoto2024steering} or to select actions that obey safety constraints~\cite{wu2025foresight}. Using a WM allows the robot to steer its policy based not just on the next action, but on a projected series of actions. We build on ideas from both planning with WMs and policy steering by learning a value function that steers the actions of a behavior-cloned policy toward states within the data distribution and closer to the task goal.

\subsection{Offline \& Inverse Reinforcement Learning}
We propose a value function that favors states in the distribution of the training data, which is related to the pessimism used in offline RL \cite{fujimoto2019off}\cite{yu2020mopo}\cite{kidambi2020morel}\cite{kumar2020conservative}, and the concept of Inverse RL \cite{ng2000algorithms}. In offline RL, a model (e.g., a value function or a dynamics model) only has access to the state-action pairs in an offline dataset. Therefore, predictions for state-action pairs outside the training distribution can be unreliable. For example, a value function could be overly optimistic when queried with out-of-distribution state-action pairs, leading to extrapolation errors \cite{fujimoto2019off}. 

\subsection{Optical Flow as a Representation in Robotics}
Both 2D optical flow and 3D point flow have been used as intermediate representations in robotics. Multiple works have used 3D flow to represent object affordances~\cite{eisner_flowbot3d_2024}\cite{zhang_flowbot_2024}\cite{yuan_general_2024}. These representations transfer well from simulation to real hardware and can be learned from human videos. Other works have predicted 2D optical flow as an action representation~\cite{lin_flowretrieval_2024}\cite{goyal_ifor_2022} or used it to measure the similarity between actions across episodes~\cite{lin_flowretrieval_2024}\cite{verghese_skills_2024}. Rather than using optical flow as a predicted action representation, in this work, we use optical flow as an embodiment-agnostic transition representation to learn a WM across diverse embodiments, with the goal of achieving efficient cross-embodiment transfer.

%% file: method.tex
\section{Preliminaries}

We consider robot control in a Partially Observable Markov Decision Process (POMDP) setting, with a latent state space learned by a neural network. A POMDP can be described by a tuple $ ( \Omega, \mathcal{X, S, A, R, P} )$, which consists of observations $x \in  \mathcal{X}$,  conditional observation probabilities $\Omega(x' | s, a)$, states $s \in \mathcal{S}$, actions $ a \in \mathcal{A}$, reward function $r = \mathcal{R}(s, a)$, and transition function $P(s' | s, a)$. For $t \in [1,2, ..., T]$, observation, state, action, and reward are denoted as $x_t, s_t, a_t, r_t$. Given the discount factor $\gamma$, the future expected discount reward from the state $s_t$ is defined as: $E[ \sum_{t=0}^{T} \gamma^t r_t ]$ and can be estimated by a state-value function $\mathcal{V}(s_t)$. Given a task horizon length $T$, policy steering seeks the best action such that: $a_t^{*}
=
\text{argmax}_{a \in \mathcal{A}}
\mathbb{E}
\left[
\sum_{k=t}^{T}
\gamma^{\,k-t} r_k
\;\middle|\;
s_t,\,
a_t = a
\right]$. 

A dataset consists of sequences of ($x_t, a_t, r_t$) where $r_t$ is a binary reward indicating task success. In this work, we consider two such datasets: a small dataset $\mathcal{E}$ of expert demonstrations on the target robot embodiment, and a larger cross-embodiment dataset $\mathcal{C}$ with action space $\mathcal{A}'$ that may not match the action space $\mathcal{A}$ of the target embodiment and sequences that do not necessarily represent task success. Importantly, $\|\mathcal{C}\| >> \|\mathcal{E}\|$, and we would like to maximally leverage $\mathcal{C}$ in training despite its action-space mismatches and non-optimal data. 

\section{Methods}

\subsection{Embodiment-Agnostic Pretraining}\label{section:flow-as-action}

We hypothesize that a WM that transfers easily to a new embodiment should depend less on embodiment-specific information. Thus, we train an image-based WM by dropping both proprioception and robot actions from the pretraining phase. Motivated by the observation that different embodiments share similar visual motion patterns when they execute similar skills (e.g., picking up an object), as shown in Fig.~\ref{fig:optical_flow_visualization}, we train the WM to model visual dynamics, given optical flow as embodiment-agnostic transition representation. 

Our WM adopts the objective and architecture from Dreamer v3 \cite{hafner2023mastering} and learns an optical flow encoder during pretraining (middle of Fig.~\ref{fig:motivation1}). Compared to previous work \cite{lin_flowretrieval_2024}\cite{bi2026motus}, which must encode optical flow with a VQ-VAE \cite{van2017neural} in a separate stage before pretraining begins and thus incurs additional training overhead, our optical flow encoder is optimized end-to-end with the WM. Its architecture is based on convolution layers and projects the feature map to an $n$-dimensional vector. We set $n=\|\mathcal{A}\|$, the dimension of the action space of the target embodiment. Since $n$ is much smaller than the dimensionality of the optical flow image, the encoder must extract the salient features of the flow and discard redundant, motion-irrelevant information (e.g., the morphology of the embodiment, spatial details of the scene). This prevents future information carried by the optical flow from leaking into the WM's predictions, which would otherwise make the pretraining objective trivially easy. 

\subsection{Policy-Agnostic Finetuning}\label{section:finetuning}

Given a small dataset from the target embodiment $\mathcal{E}$, the pretrained WM is finetuned to use the action space of the target embodiment. This finetuning also makes the WM capable of steering a policy towards better actions on the target embodiment, agnostic to the policy architecture. To this end, we assume that a base policy $\pi$ is learned on the target embodiment data (upper right of Fig.~\ref{fig:motivation1}), producing a distribution over actions for any observation $a_t \sim \pi(x_t)$. Then, we replace the optical flow encoder of the WM, which projects optical flow to an $\|\mathcal{A}\|$-dimensional vector, with normalized robot actions in this same $\|\mathcal{A}\|$-dimensional space (middle of Fig.~\ref{fig:motivation1}). When finetuning the WM on $\mathcal{E}$, we again use the Dreamer v3 \cite{hafner2023mastering} objective. Although there is naturally a distribution shift between the compressed optical flow representation and normalized robot actions, we found that the WM can efficiently adapt to the robot action space after finetuning its parameters with only 50 demonstrations.

To steer a policy towards better actions, we train a robust value function $\mathcal{V}(s_t)$ with the WM during finetuning, which estimates discounted future rewards. This value function leverages the learned latent state $s_t$ from the WM, but its gradients do not flow back into the WM. Crucially, our value function must be able to estimate discounted future rewards across states the policy is likely to visit during inference, not just states in the expert dataset. Furthermore, the base policy may become unstable if the robot deviates too far from states in the expert dataset, so our value function should also ensure the robot stays close to states in that expert dataset. 

To train a value function with both of these qualities, we adopt the following procedure. First, we assume that a base policy has been trained on the target embodiment dataset $\mathcal{E}$. A predicted action plan \footnote{An action plan is a sequence of actions predicted jointly by the policy and executed open-loop. This technique is often referred to as action chunking \cite{chi2023diffusion}\cite{zhao2023learning} and leads to more robust performance.\label{fn:action_plan}} is sampled from the policy (line 4, Alg.~\ref{alg:lps_single}). The resulting latent states of the predicted action plan are compared against the latent states from the expert dataset via a similarity metric, and the similarity is converted into an additional reward penalizing distribution shifts (lines 6 and 8, Alg.~\ref{alg:lps_single}). We design the reward around cosine similarity because its bounded range helps stabilize training of the value function, and because of its prevalent use in the deep learning literature \cite{radford2021learning}\cite{chen2021exploring}. Lastly, a robust value function is obtained by training on both states likely to be visited by the policy during inference and states from the expert dataset (lines 10-11, Alg.~\ref{alg:lps_single}), using a lambda-return objective adopted from Dreamer v3 \cite{hafner2023mastering}.

\begin{algorithm}[ht]
\caption{Robust value function for LPS}\label{alg:lps_single}
\begin{algorithmic}[1]
\State \textbf{Requires:} Robot dataset $\mathcal{E}=[ (x_{1:T}, a_{1:T}, r_{1:T})_n | n=[1, ..., N]]$, world model $\mathcal{W}$, policy $\pi$, similarity metric $sim$, lambda-return computation $r_{\lambda}$, and loss $\mathcal{L}$. \\
\textbf{Initialize:} the value function $\mathcal{V}$, $\mathcal{V}_{target}$, $h$, $\gamma$, and $\lambda$.

\State Sample $(x_{t:t+h}, a_{t:t+h}, r_{t:t+h})$ from $\mathcal{E}$. 

\State Sample a predicted action plan\footref{fn:action_plan}: $ a'_{t:t+h} \sim \pi( x_{t})$.
\State $s_t = \mathcal{W}.\text{enc}(x_t)$ \Comment{  observations to latent states.}
\State $s_{t:t+h}, s'_{t:t+h} = \mathcal{W}(s_t, a_{t:t+h}), \mathcal{W}(s_t, a'_{t:t+h})$ 
\State $v_{t:t+h}, v'_{t:t+h} = \mathcal{V}_{target}(s_{t:t+h}), \mathcal{V}_{target}(s'_{t:t+h})$ 
\State $r'_{t: t + h} = r_{t: t + h} +$ $(sim(s_{t: t + h}, s'_{t: t + h})-1)/2$
 \hspace{0.5cm} \Comment{ reward penalizing deviations from the expert data}
\State $R_{\lambda}, R'_{\lambda}= r_{\lambda}(r_{t:t+h}, v_{t:t+h}, \gamma, \lambda ),r_{\lambda}(r'_{t:t+h}, v'_{t:t+h}, \gamma, \lambda )$
\State $R_{\text{all}}, S_{\text{all}} = \text{concat}(R_{\lambda}, R'_{\lambda}), \text{concat}(s_{t:t+h}, s'_{t:t+h}) $
\State Optimize the value function: $\mathcal{L}( R_{\text{all}}, \mathcal{V}(S_{\text{all}}))$.
\State Update $\mathcal{V}_{target}$ from $\mathcal{V}$.
\end{algorithmic}
\end{algorithm}

\subsection{Inference via Latent Policy Steering}
 LPS improves the base policy during inference by evaluating multiple candidates in the latent state space of the WM and selecting the best one according to the value function. We sample $B$ action plans \footref{fn:action_plan} with horizon $h$. After encoding the current observation, the WM predicts the future states for each action plan with the state values. A plan-level value is computed as a discounted weighted average of the predicted state values, assigning heavier weights to later states to reduce noise in the value function's predictions. In practice, we find that this discounted weighted average works better than using the unweighted mean or the final state value to represent a plan's value. Finally, the action plan \footref{fn:action_plan} with the highest value is executed in open-loop for the given horizon before replanning.

%% file: experiment.tex
\section{Experiments}

\begin{figure}[t]
      \centering
     
      \includegraphics[scale=0.95]{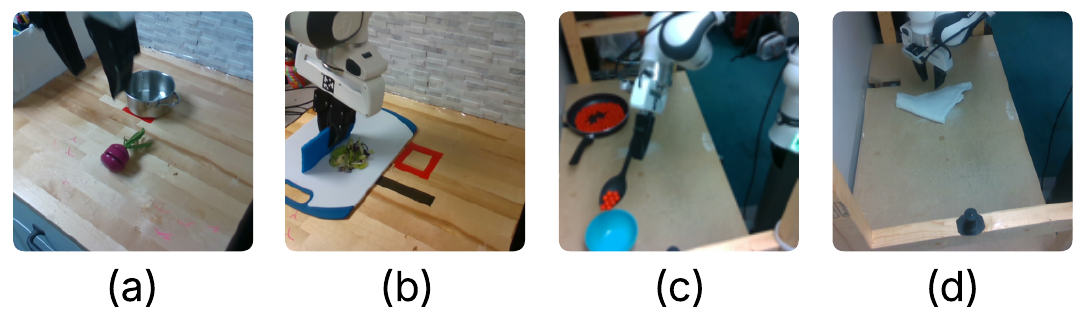}
      \caption{ 
      \textbf{Real-world experiment}. The proposed method and baselines are evaluated on: (a) Put-radish-in-pot, (b) Sweep-salad-off-the-board, (c) Scoop-beads-with-spoon, and (d) Fold-towel-to-triangle. }
      \vspace{-1.8em}
      \label{fig:realworld}
   \end{figure}

We evaluate the proposed method on real-world environments and on a simulated benchmark (Robomimic \cite{robomimic2021}), with a strict cross-embodiment transfer setup. Specifically, the \textbf{Franka robot} is chosen to be the target embodiment for finetuning and evaluations, and \textbf{is removed from the pretraining phase}, as shown in Fig.~\ref{fig:data}. In the following, we describe our environments for evaluations, pretraining/finetuning practices, and baseline implementations. 

\subsection{Settings}

\textbf{Environment: Robomimic.} Our approach and baselines are evaluated on the Franka robot in 4 manipulation tasks (Lift, Can, Square, Transport) from Robomimic~\cite{robomimic2021}. While Lift, Can, and Square involve only single-arm pick-and-place skills, Transport requires bimanual, long-horizon manipulation, allowing us to assess the proposed method's effectiveness in a more challenging setting. For evaluation, we report average success rates and standard deviations across 3 seeds, where each seed's success rate is computed over 150 episodes.

\textbf{Environment: Real world.} The environment includes a Franka robot, a desk, and objects to manipulate. We mount a camera on the robot's wrist and a fixed camera on the side. The target dataset is collected via teleoperation with a joystick. Objects positions during data collection are randomly selected from the desk area. For evaluation, we use 20 pre-selected object positions that uniformly cover the desk area. We consider 4 manipulation tasks that include pick-and-place, tool use, or deformable objects, shown in Fig.~\ref{fig:realworld}. Scooping requires the robot to use a spoon to move beads from a pan to a bowl; imprecise manipulation easily causes beads to fall out of the spoon. Folding and sweeping require manipulating deformable objects.

\begin{figure}[t]
      \centering
     
      \includegraphics[scale=0.35]{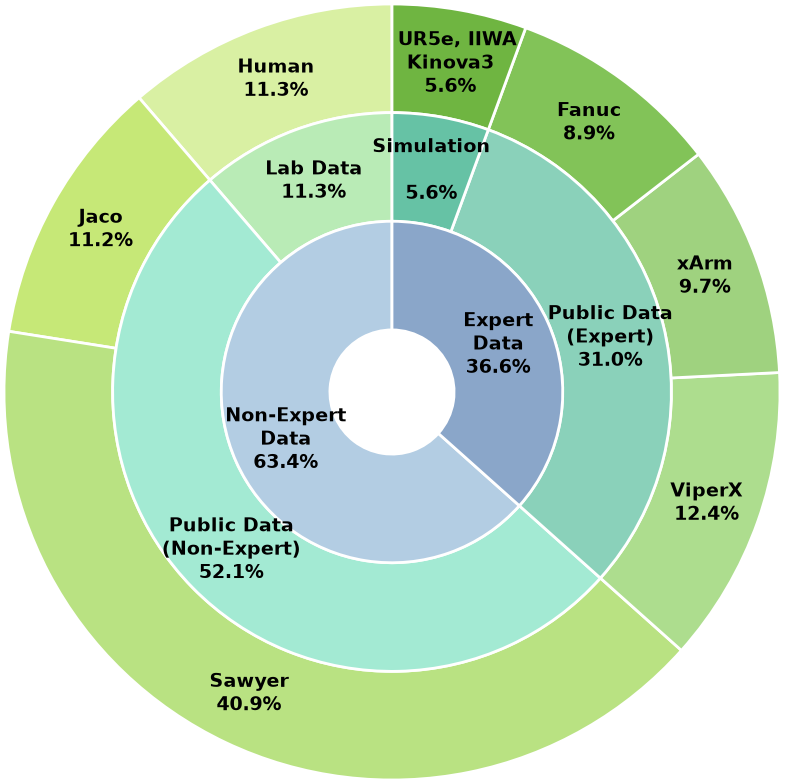}
      \caption{ 
      \textbf{Pretraining data curation}. This chart summarizes the pretraining data by quality (inner ring), domain (middle ring), and embodiment (outer ring). The dataset comprises roughly 55 hours of interaction data across 9 embodiments from different domains, including non-expert data.}
      \vspace{-2.0em}
      \label{fig:data}
   \end{figure}

\textbf{Pretraining.} To strictly evaluate the proposed framework for cross-embodiment transfer, we curate the pretraining dataset $\mathcal{C}$ without the target embodiment (i.e., the Franka robot), drawing on public dataset \cite{vuong2023open} or cost-effective data sources that are sub-optimal (e.g., human play videos). This results in roughly 55 hours of data and 15K episodes. While human data for pretraining is collected in the same environment used for real-world evaluations, the human is instructed to play with objects (including both the target and irrelevant objects) instead of demonstrating the task. For simulation data, we use Robosuite~\cite{robosuite2020} with a joystick to collect 100 demonstrations in 3 tasks (Lift, Can, Square), each with 3 robots other than the Franka: IIWA, UR5e, and Kinova3, for a total of 900 demonstrations. An overview of the data composition is shown in Fig.~\ref{fig:data}. The optical flow used for pretraining is computed between two consecutive video frames via GMFlow \cite{xu2022gmflow}. 
   
\textbf{Finetuning.} The expert dataset $\mathcal{E}$ is collected on the target-embodiment Franka robot. Both the Robomimic and real-world evaluations use only 50 demonstrations, testing the efficiency of the proposed framework. 

\textbf{Baselines.} We compare against baselines learned from scratch and baselines pretrained across embodiments and then finetuned on the target embodiment. All policies (DP, HPT, and Pi0.5) use an action chunking horizon of 16 \footref{fn:action_plan}. The prediction horizon of LPS's WM is also set to be 16.
\begin{itemize}
    \item DP: a Diffusion Policy \cite{chi2023diffusion} with 550K parameters learned by behavior cloning on the target dataset $\mathcal{E}$.
    \item LPS-DP-scratch: the WM has 33M parameters and is learned from scratch on the target dataset $\mathcal{E}$. Its base policy is the DP baseline.
    \item HPT: a cross-embodiment pretrained policy \cite{wang2024scaling} adopting a transformer trunk to extract shared representations across embodiments with embodiment-specific action heads and encoders. We choose to finetune the HPT-Base variants \cite{wang2024scaling} with 12.6M parameters.
    \item Pi0.5: an open-weight VLA \cite{intelligence2025pi} with 3.6B parameters, pretrained on various embodiments and finetuned on the target dataset $\mathcal{E}$. By applying LoRA finetuning \cite{hu2022lora}, it has approximately 30M trainable parameters.
    \item LPS-(DP/Pi0.5): the WM has 33M parameters, pretrained on the multi-embodiment dataset $\mathcal{C}$, and finetuned on the target dataset $\mathcal{E}$. To evaluate the proposed policy-agnostic finetuning, we choose to combine the WM with the DP or Pi0.5 baselines.
\end{itemize} 

\textbf{Inference details for LPS.} An RTX 3090 with 24 GB of VRAM is used for evaluations. With DP as the base policy, LPS uses a sample size $B=400$ (the sampling overhead is negligible for a lightweight policy like DP). Evaluating 400 candidate plans with the WM adds roughly 100 ms of latency. Since Pi0.5 has 3.6B parameters, we set the sample size $B=16$ for LPS-Pi0.5 due to memory constraints.

 \begin{figure*}[t]
      \centering
      \includegraphics[scale=0.18]{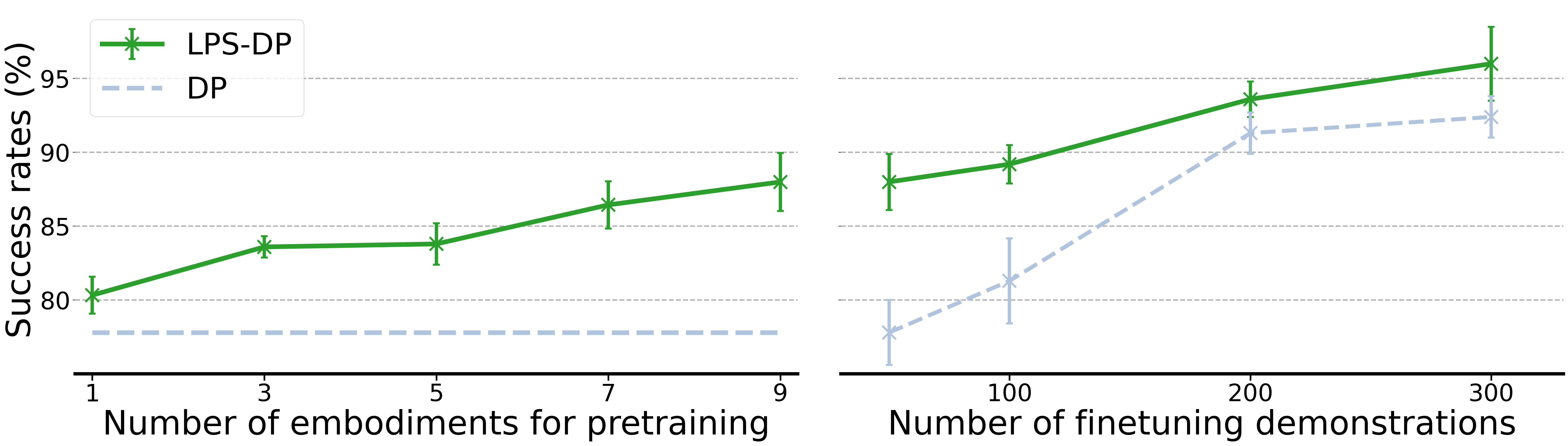}
      \caption{ 
      \textbf{Scalability and sample efficiency}. Left: LPS's WM is pretrained on an increasing number of embodiments, and finetuned with 50 demonstrations. Thanks to the embodiment-agnostic pretraining, LPS's performance scales effectively when provided with more data from diverse embodiments during pretraining.
      Right: LPS's WM is pretrained on all embodiments and then receives increasing numbers of demonstrations (50--300) to finetune the WM and train the base policy. It takes at least \textbf{twice the data for the DP baseline to outperform LPS-DP} (LPS-DP with 50 demonstrations outperforms DP with 100 demonstrations), highlighting the sample efficiency of LPS in the low-data regime.  }
      \vspace{-1.5em}
      \label{fig:scaling}
   \end{figure*}

\subsection{Evaluations}

\input{robomimic}

\input{real_world}  
\textbf{ Does LPS effectively transfer multi-embodiment data to an unseen target embodiment, agnostic to the chosen base policy?} Given only 50 demonstrations collected on an unseen embodiment, LPS effectively improves the chosen base policy, including a lightweight policy learned from scratch (DP) and a finetuned VLA (Pi0.5), consistently across tasks in simulation (Table~\ref{table:robomimic}) and real-world evaluations (Table~\ref{table:realworld}). This result demonstrates both the efficiency of LPS, enabled by embodiment-agnostic pretraining, and its flexibility, enabled by policy-agnostic finetuning. We find that LPS-DP performs best in simulation (Table~\ref{table:robomimic}), while LPS-Pi0.5 performs best in the real world (Table~\ref{table:realworld}) — likely because Pi0.5's pretraining dataset is largely collected in the real world, making it a stronger base policy when finetuned with LPS for real world tasks.


Although both HPT and Pi0.5 are pretrained on large multi-embodiment datasets, Pi0.5 significantly outperforms HPT, likely due to its far greater capacity (3.6B vs. 12.6M parameters), which allows it to learn more transferable representations from diverse embodiments. We therefore exclude HPT from the real-world evaluation. 

\subsection{Ablations}

In our ablation studies, we use the Can task from the Robomimic benchmark to examine the design choices and properties of the proposed framework. 

\textbf{
How does LPS’s performance scale with different amounts of pretraining and finetuning data?
} We investigate LPS's scalability with increasing embodiment gaps by pretraining the WM on varying amounts of data spanning different numbers of embodiments (Fig.~\ref{fig:scaling}, left), then finetuning it with 50 target-embodiment demonstrations and combining it with a base policy trained from scratch on the same data. Thanks to embodiment-agnostic pretraining, LPS's performance scales with pretraining dataset size despite the growing embodiment gaps.

To further understand the sample efficiency of LPS, the WM pretrained on all embodiments is finetuned with 50 to 300 target-embodiment demonstrations and compared to a base policy (DP) trained on the same amount of data (Fig.~\ref{fig:scaling}, right). The performance gain resulting from LPS is significant in the low-data regime: DP with 100 demonstrations still does not outperform LPS with access to only 50 demonstrations, highlighting the sample efficiency of LPS. The gain from LPS is smaller given sufficient data (200--300 demonstrations), potentially due to performance saturation. 

\textbf{
How does pretraining, robust value function, and reward designs contribute to LPS’s performance?} Given 50 demonstrations on the target embodiment, we conduct the following ablations based on LPS-DP.

\begin{itemize}

    \item no-pretrain-no-robust: the WM is learned from scratch. To isolate the effect of training the value function $\mathcal{V}$ on additional non-expert states $s'_{t:t+h}$ (which improve robustness) together with the rewards $r'_{t:t+h}$ (which help avoid distribution shift), both are removed from the value function training (i.e., line 10, Alg.~\ref{alg:lps_single} changes to $R_{\text{all}}, S_{\text{all}} = R_{\lambda}, s_{t:t+h}$). 
    \item no-pretrain-no-reward: the WM is learned from scratch. To isolate the effect of training the value function $\mathcal{V}$ on additional non-expert states without the negative rewards $r'_{t:t+h}$ that penalize distribution shift, the additional rewards are removed from the objective (i.e., line 8, Alg.~\ref{alg:lps_single} changes to $r'_{t:t+h} = r_{t:t+h}$). 
    \item no-pretrain: this is LPS-DP-scratch, i.e., a WM learned from scratch on the target embodiment.
\end{itemize}

\begin{figure}[t]
      \centering
     
      \includegraphics[scale=0.16]{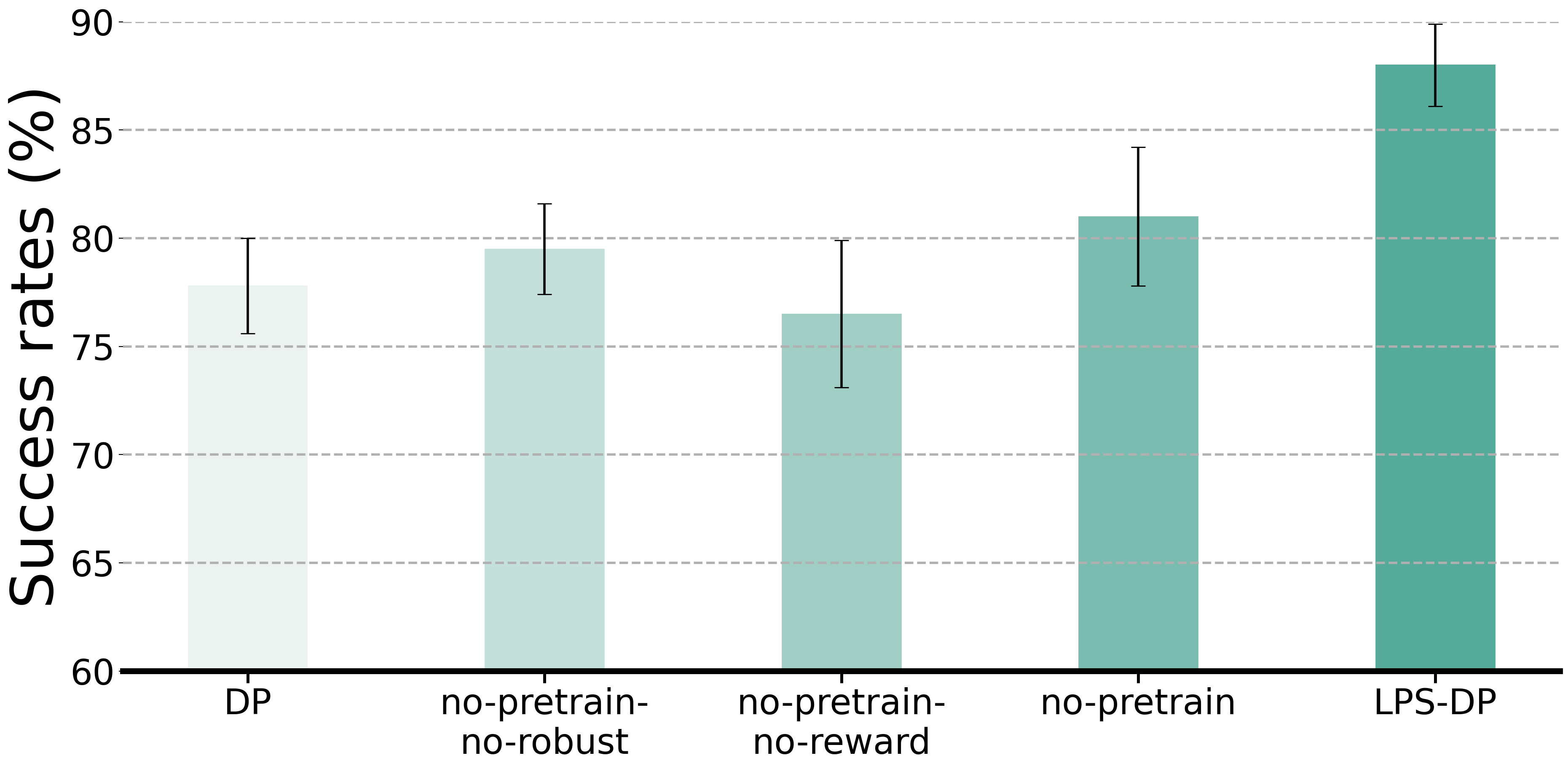}
      \caption{ 
      \textbf{Ablations on LPS's design}. Pretraining the WM brings the largest improvement in this low-data regime (LPS-DP vs. no-pretrain). In addition, naively learning a value function for policy steering from expert states with binary rewards (no-pretrain-no-robust) provides only marginal benefit, and learning from extra states without the reward that penalizes distribution shift decreases performance (no-pretrain-no-reward). }
      \vspace{-1.5em}
      \label{fig:ablations}
   \end{figure}

As shown in Fig.~\ref{fig:ablations}, the largest gain comes from the pretrained WM (LPS-DP vs. no-pretrain, 88.0\% vs. 81.0\%). Since LPS relies on a value function to steer the policy toward better actions in the WM's learned state space, a poorly learned state space limits its effectiveness; pretraining mitigates this by providing a better-structured state space for finetuning. Further, we find that naively learning a value function from expert states with binary rewards does not meaningfully improve over the base policy (no-pretrain-no-robust vs. DP, 79.5\% vs. 77.8\%), while training the value function on states beyond the dataset without the reward that penalizes distribution shift degrades performance below the base policy itself (no-pretrain-no-reward, 76.5\%). Thus, for effective policy steering, it is crucial to consider both states that are sampled from the dataset ($s_{t:t+h}$) and states that are likely to be visited during inference ($s'_{t:t+h}$), with a preference to stay close to the dataset by penalizing states far from it ($r'_{t:t+h}$).

\textbf{How does horizon length affect LPS’s performance?} We compare LPS-DP-scratch with DP given 100 demonstrations on the Can task of Robomimic with different horizon lengths. The prediction horizon of the WM always matches the action chunking horizon \footref{fn:action_plan} of DP. LPS-DP-scratch outperforms DP on horizons 4, 8, 12, and 16, and performs worse on a horizon of 24 (Fig.~\ref{fig:lps_horizon}). A large action prediction size leads to a suboptimal policy, causing the reward used to penalize distribution shift to become noisy (line 8, Alg.~\ref{alg:lps_single}). As a result, the value function for policy steering becomes less effective.

%% file: robomimic.tex
\begin{table}[t]
\caption{Latent Policy Steering in Robomimic}

\label{table:robomimic}
\setlength{\tabcolsep}{3.5pt}
\begin{tabular}{@{}lllllll@{}}
\toprule
\multirow{2}{*}{\begin{tabular}[c]{@{}l@{}}Task \textbackslash\\ Setting\end{tabular}} &
  \multicolumn{2}{c}{From scratch} &
  \multicolumn{4}{c}{Pretrain-and-finetune} \\

\cmidrule(lr){2-3} \cmidrule(lr){4-7}
 &
  DP &
  \begin{tabular}[c]{@{}c@{}}LPS-DP-\\ scratch\end{tabular} &
  HPT &
  Pi0.5 &
  \begin{tabular}[c]{@{}c@{}}LPS-\\ DP\end{tabular} &
  \begin{tabular}[c]{@{}c@{}}LPS-\\ Pi0.5\end{tabular} \\ 
Lift      & 83.9±1.5             & 86.9±1.7              & 58.2±2.6   & \textbf{100}±0.0         & 90.1±1.8          & \textbf{100}±0.0     \\
Can       & 77.8±2.2        & 81.0±3.2       & 79.4±3.5   & 87.8±0.8   & 88.0±1.9   & \textbf{94.4}±1.1  \\
Square    & 40.5±3.8        & 46.3±2.7       & 32.5±2.5   & 30.7±1.0          & \textbf{53.4}±4.8   & 35.1±1.2        \\
Transport & 26.4±1.8               & 29.5±3.2              & 11.5±1.4   & 25.1±4.6          & \textbf{35.3}±3.0          & 34.5±2.0         \\  \midrule
Average   & 57.2\%               & 60.9\%              & 45.4\% & 60.9\%          & \textbf{66.7}\%          & 66.0\%         \\ \bottomrule
\end{tabular}
\vspace{0.5em}
\\We report the mean success rate across 3 seeds with standard deviations; each seed has 150 episodes of evaluation. With only 50 demonstrations on the target embodiment, LPS consistently improves both base policies, the Diffusion Policy (DP) and the state-of-the-art VLA (Pi0.5), across all tasks, demonstrating its sample efficiency for cross-embodiment transfer and its flexibility in accommodating different base policies.
\vspace{-1.5em}
\end{table}

%% file: real_world.tex
\begin{table}[b]

\caption{Latent Policy Steering in the real world}
\label{table:realworld}
\vspace{-0.5em}
\begin{center}
\begin{tabular}{@{}llllll@{}}
\toprule
\multirow{2}{*}{Task\textbackslash{}Settings} &
  \multicolumn{2}{c}{From scratch} &
  \multicolumn{3}{c}{Pretrain-and-finetune} \\

\cmidrule(lr){2-3} \cmidrule(lr){4-6}
 &
  DP &
  \multicolumn{1}{c}{\begin{tabular}[c]{@{}c@{}}LPS-DP-\\ scratch\end{tabular}} &
  Pi0.5 &
  \multicolumn{1}{c}{\begin{tabular}[c]{@{}c@{}}LPS-\\ DP\end{tabular}} &
  \multicolumn{1}{c}{\begin{tabular}[c]{@{}c@{}}LPS-\\ Pi0.5\end{tabular}} \\ 
\begin{tabular}[c]{@{}l@{}}Put-radish-\\ in-pot\end{tabular}         & 12/20           & 13/20          & \textbf{20}/20           & 18/20          & \textbf{20}/20          \\
\begin{tabular}[c]{@{}l@{}}Sweep-salad-\\ off-the-board\end{tabular} & 5/20            & 7/20           & 10/20           & 11/20          & \textbf{12}/20         \\
\begin{tabular}[c]{@{}l@{}}Scoop-beads-\\ with-spoon\end{tabular}    & 13/20               & 15/20              & 16/20           & 18/20              & \textbf{20}/20          \\
\begin{tabular}[c]{@{}l@{}}Fold-towel-\\ to-triangle\end{tabular}    & 7/20               & 7/20              & 9/20           & \textbf{13}/20              & 12/20          \\  \midrule
Average                                                              & 46.2\%               & 52.5\%              & 68.8\%           & 75.0\%              & \textbf{80.0}\%          \\ \bottomrule
\end{tabular}
\end{center}
We report the number of successes out of 20 trials in the real world. Based on 50 demonstrations on the target embodiment, LPS improves both base policies, the Diffusion Policy (DP) and the finetuned VLA (Pi0.5), across all tasks, demonstrating both its efficiency and its flexibility for cross-embodiment transfer.
\end{table}